\documentclass[twoside,11pt]{article}

\usepackage{jmlr2e}

\usepackage{cite}
\usepackage{amsmath,amssymb,amsfonts}
\usepackage{algorithm}
\usepackage{algorithmic}
\usepackage{graphicx}
\usepackage{textcomp}
\usepackage{xcolor}
\usepackage{hyperref}
\usepackage{multirow}
\usepackage{siunitx}                 
\usepackage{booktabs, threeparttable}
 
\usepackage[inline]{enumitem}
\usepackage{fdsymbol}

\usepackage{caption} 
\usepackage[symbol]{footmisc}

\DeclareMathOperator{\argmax}{argmax}

\DeclareMathOperator{\opt}{opt}

\newcommand{\calN}{\mathcal{N}}

\newcommand{\bfW}{\mathbf{W}}
\newcommand{\bfX}{\mathbf{X}}

\newcommand{\bfy}{\mathbf{y}}

\jmlrheading{1}{}{}{}{}{}

\firstpageno{1}

\begin{document}

\title{Bayesian Optimization for Macro Placement}

\author{\name Changyong Oh$^{1,}$\thanks{~~Work completed during internship at Qualcomm Technologies Netherlands B.V. Qualcomm AI Research is an initiative of Qualcomm Technologies, Inc.} \email changyong.oh0224@gmail.com
       \AND
       \name Roberto Bondesan$^2$ \email rbondesa@qti.qualcomm.com
       \AND
       \name Dana Kianfar$^2$ \email dkianfar@qti.qualcomm.com
       \AND
       \name Rehan Ahmed$^3$ \email rehaahme@qti.qualcomm.com
       \AND
       \name Rishubh Khurana$^4$ \email rishkhur@qti.qualcomm.com
       \AND
       \name Payal Agarwal$^4$ \email payaagar@qti.qualcomm.com
       \AND
       \name Romain Lepert$^2$ \email romain@qti.qualcomm.com
       \AND
       \name Mysore Sriram$^4$ \email mysoresr@qti.qualcomm.com
       \AND
       \name Max Welling$^1$ \email m.welling@uva.nl
       \AND
       \addr $^1$ QUvA lab, University of Amsterdam, Netherlands\\
       \addr $^2$ Qualcomm AI Research, Netherlands\\
       \addr $^3$ QT Technologies Ireland Limited\\
       \addr $^4$ QUALCOMM India Private Limited\\
       }

\maketitle

Macro placement is the problem of placing memory blocks on a chip canvas. 
It can be formulated as a combinatorial optimization problem over sequence pairs, a representation which describes the relative positions of macros.
Solving this problem is particularly challenging since the objective function is expensive to evaluate.
In this paper, we develop a novel approach to macro placement using Bayesian optimization~(BO) over sequence pairs.
BO is a machine learning technique that uses a probabilistic surrogate model and an acquisition function that balances exploration and exploitation to efficiently optimize a black-box objective function. 
BO is more sample-efficient than reinforcement learning and therefore can be used with more realistic objectives. 
Additionally, the ability to learn from data and adapt the algorithm to the objective function makes BO an appealing alternative to other black-box optimization methods such as simulated annealing, which relies on problem-dependent heuristics and parameter-tuning.
We benchmark our algorithm on the fixed-outline macro placement problem with the half-perimeter wire length objective and demonstrate competitive performance.

\begin{keywords}
  Bayesian optimization, Permutation, Batch acquisition, Physical design, Macro Placement, Sequence Pair
\end{keywords}

\section{Introduction}

In chip placement two different types of objects are placed on a chip canvas: macros, which are large memory blocks, and standard cells, which are small gates performing logical operations.
Compared to macros, standard cells are typically thousands of times smaller but tens or hundreds of thousands of times more numerous.
While standard cell placement can be efficiently solved using continuous optimization, e.g.~\citep{replace}, macro placement is typically framed as a combinatorial optimization problem due to their larger physical size. This involves searching over the discrete set of relative positions between pairs of macros, e.g.~whether macro $i$ is to the left or right of macro $j$, which act as constraints against overlapping macros.
The most popular combinatorial representation of relative positions is called the sequence pair which is composed of a pair of permutations, one per spatial dimension~\citep{murata1996vlsi}.

The goal of macro placement is to place macros in such a way that the power, performance and area metrics are jointly optimized.
The combinatorial nature and varying sizes of macros and standard cells, together with the cost of evaluating the objective function (several days for complex designs), make macro placement a notoriously challenging step in physical design. 
Macro placement is also related to floorplanning, where standard cells are clustered in soft rectangles that are jointly placed with hard rectangles that represent macros \citep{kahng2011vlsi}.
In practice, designers manually place macros based on their intuition which is likely sub-optimal.

Machine learning  algorithms offer an advantage over traditional optimization algorithms for macro placement since they can learn from past designs and improve over time in an automated fashion, adapting the algorithms to specific use cases.
Applying machine learning to physical design has therefore recently emerged as a main research effort in electronic design automation~\citep{kahng2018machine}.
In particular, reinforcement learning (RL) provides a natural framework for automating design decisions, where an agent plays the role of a designer in carefully selecting parameter configurations to evaluate next while searching for optimal solutions. However, in practice applying RL is very costly because of the large number of samples required for learning a good policy, due in part to the very large design space and costly evaluation as remarked above.
For this reason, to the best of our knowledge, applications of RL in the literature are either limited to a handful of parameters \citep{Agnesina2020} or require the use of cheap proxies instead of the real objective \citep{mirhoseini2020chip}, which changes the focus towards designing good proxies. 

Bayesian optimization (BO) is a technique that is well-known for its sample-efficiency, whereby it carefully explores the optimization landscape through selecting a candidate based on previous evaluations~\citep{shahriari2015taking}.
Compared with other black-box function optimization methods such as RL, genetic algorithms, or simulated annealing (SA), the sample efficiency of BO allows flexibility for the macro placement.
Especially when it is desirable to perform optimization close to the real objective, not a proxy, inevitable evaluation cost leaves only BO as a viable option.
For the application to macro placement, more relevant is BO on combinatorial structures~\citep{baptista2018bayesian,oh2019combinatorial,deshwal2021mercer,deshwal2021combining,oh2021batch,deshwal2022bayesian}.
For the details on BO on combinatorial structures, please refer to the references.

\begin{figure*}[tb!]
    \centering
    \includegraphics[width=.9\textwidth]{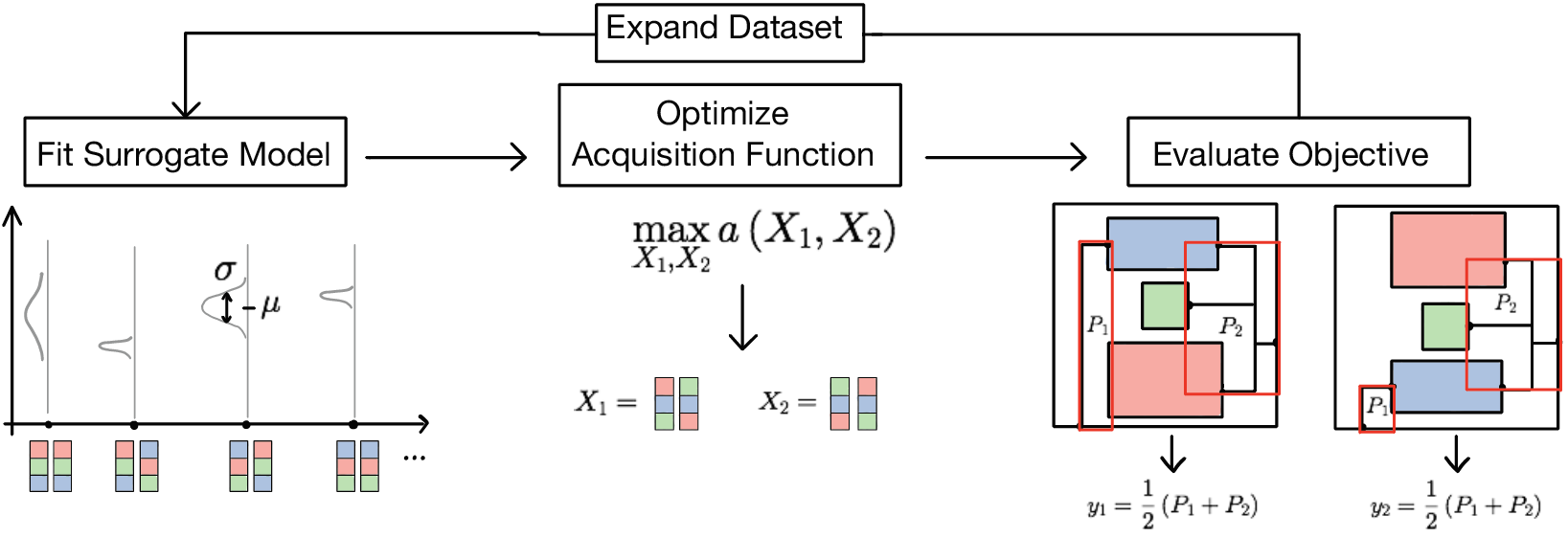}
    \caption{The Bayesian optimization for macro placement workflow for $N=3$ macros and batch size $B=2$. 
    First we fit a surrogate model (Gaussian process) to the data, updating the mean $\mu$ and standard deviation $\sigma$ of the cost function estimate for each sequence pair, here represented as a pair of $N$ dimensional arrays where permutations correspond to different colors patterns.
    Then we optimize an acquisition function $a$  conditioned on the data observed so far to find new sequence pairs $x_1, \dots, x_B$. Next we evaluate $x_1, \dots, x_B$ by placing macros to minimize HPWL while respecting the sequence pair constraints, and compute the corresponding objective values $y_1, \dots, y_B$. Here $P_i$ is the perimeter of the $i$-th bounding box of the net between macros and I/O pads.
    Finally we add these new points to the dataset and repeat the procedure until the computational budget is exhausted.
    }
    \label{fig:bo_flow}
\end{figure*}

{\bf Contributions} In this paper we introduce BO on sequence pairs for macro placement as a replacement for other black-box optimization methods such as SA which are routinely applied in the literature~\citep{adya2002consistent, 10.1145/1044111.1044116}.
We use batch BO for parallel evaluation of a batch of data points to accelerate the optimization.
Fig.~\ref{fig:bo_flow} summarizes our workflow.
Our main contributions are:
\begin{itemize}
    \setlength\itemsep{0em}
    \item We extend batch BO on permutations \citep{oh2021batch} to batch BO on sequence pairs and devise an efficient algorithm for parallel batch acquisition function optimisation. 
    \item We benchmark our algorithm on the MCNC dataset and
    obtain superior performance in terms of wire length metric as compared to SA.
\end{itemize}

\section{Methodology}\label{sec:method}
In contrast to the traditional macro placement approaches, we consider an expensive-to-evaluate objective with which we perform macro placement.
We also consider the fixed-outline constraint  addressed in many existing works~\citep{adya2001}.

In order to efficiently tackle macro placement, we employ batch BO on the space of sequence pairs~(SPs) -- a pair of two permutations -- a compact representation for the relative positions of macros.
Sequence pairs describe non-overlapping placements of macros.
Intuitively, if we imagine the macros to be placed on a line, the space of non-overlapping placements can be indexed by permutations of the macros, and a macro optimization problem with non-overlapping constraints can be solved by searching over the space of permutations.
In the two-dimensional setting of this paper we need a pair of permutation to describe non-overlapping macro placements.  
See Appx. Subsec.~\ref{subsec:sp} and~\citep{murata1996vlsi} for detailed explanation.
To this end, we introduce \textit{1)} a kernel on the space of SPs, \textit{2)} an efficient heuristic to optimize the batch acquisition function \textit{3)} an efficient vectorized Python implementation of the least common subsequence~(LCS) algorithm of $O(N \log N)$ run-time complexity where $N$ is the number of macros.

For our Gaussian process surrogate model in BO, the proposed kernel is based on the position kernel on permutation due to its superior performance in (batch) BO on permutation spaces\citep{zaefferer2014distance,oh2021batch}. 
Denoted $\pi_1,\pi_2,\pi_1',\pi_2'$ four permutations of $N$ elements, our kernel is:
\begin{equation}
    K_{sp}((\pi_1, \pi'_1), (\pi_2, \pi'_2) \vert \bfW, \bfW') = K_{perm}(\pi_1, \pi_2 \vert \bfW) \cdot K_{perm}(\pi'_1, \pi'_2 \vert \bfW') 
\end{equation}
{\setlength{\parindent}{0cm}
where $\bfW = [w_1, \cdots, w_N]$, $\bfW' = [w'_1, \cdots, w'_N]$, and
}
\begin{equation}
    K_{perm}(\pi_1, \pi_2 \vert w_1, \cdots, w_N) = \exp{\Big(\sum_{n=1}^N} w_n \cdot \vert \pi_1^{-1}(n) - \pi_2^{-1}(n) \vert \Big) \nonumber
\end{equation}

{\setlength{\parindent}{0cm}
$K_{perm}(\pi_1, \pi_2 \vert 1, \cdots, 1)$ is the position kernel~\citep{zaefferer2014distance,oh2021batch}.
In addition to the mentioned works, in our position kernel we introduce parameters $\bfW, \bfW'$ that account for widths and heights of different macros.
We optimize those parameters by maximizing the marginal likelihood of the training data by gradient descent.
}

In the batch acquisition function of BO, we adopt the method in \citep{oh2021batch} which uses determinantal point processes~(DPPs) with a weighted kernel to obtain a batch of diverse points each of which is likely to speed up BO progress.
DPPs quantify the diversity of points using determinant of the gram matrix.
Since the determinant of a matrix is the volume of hyper-parallelepiped whose vertices are columns of the matrix, the more diverse the points are, the larger the determinant is~\citep{kulesza2011k,kulesza2012determinantal}.
The batch acquisition function is defined as
\begin{equation}
    a^{(t)}_{batch}(x_1, \cdots, x_B) = \det \big[\rho\big(a^{(t)}(x_b)\big) \, K^{(t)}(x_b, x_c) \, \rho\big(a^{(t)}(x_c)\big)\big]_{b,c=1, \cdots, B} \,
\end{equation}

{\setlength{\parindent}{0cm}
where $K^{(t)}$ the covariance function conditioned on the evaluation dataset $\bfX^{(t)}$ as in Eq.~\eqref{eq:Kt}, $a^{(t)}$ is the acquisition function for a single point as in Eq.~\eqref{eq:a_single}~(e.g. EI, UCB, EST\citep{wang2016optimization}), and $\rho(\cdot)$ is a positive and strictly increasing function.
This batch acquisition function balances the quality of each point~(i.e. the likelihood of improving the objective) through the acquisition weight $\rho\big(a^{(t)}(\cdot)\big)$, and the diversity among points~(i.e. avoiding information redundancy in parallel evaluations) through $K^{(t)}$.
This function has been demonstrated to perform well for BO on permutation spaces\citep{oh2021batch}.
}

While $a^{(t)}_{batch}$ effectively fulfills the quality and diversity requirements, its optimization is computationally demanding.
In \citep{oh2021batch}, a greedy approach was employed with certain optimality guarantees. 
However, that method optimizes $a^{(t)}_{batch}$ sequentially over the batch index and limits the scalability of batch BO.
Therefore, we propose a new heuristic for parallel optimization of the batch acquisition function~(See Alg.~\ref{algo:parallel_heuristic} in Appx. Sec.~\ref{sec:algorithms})

The main idea is to perform small local updates in parallel for each element of the batch. 
Specifically, we first compute $x_{opt,1}$, the optimum of the single point acquisition function~(See line no.~1 of Alg.~\ref{algo:parallel_heuristic} in Appx. Sec.~\ref{sec:algorithms}).
Then we optimize the function $a_{s,b}$ defined by fixing all but the $b$-th element of the batch, for $b=2,\dots, B$~(See line no.~6 of Alg.~\ref{algo:parallel_heuristic} in Appx. Sec.~\ref{sec:algorithms} Alg.~\ref{algo:parallel_heuristic}). 
This step can be parallelized over the batch.
Here $s$ denotes the iteration time over which this procedure is repeated.
When a single point is updated~(line no.~1 of Alg.~\ref{algo:parallel_heuristic} in Appx. Sec.~\ref{sec:algorithms}), we apply a small local update instead of running until convergence to minimize the deviation of our individual updates from the simultaneous update method.
Intuitively, if any single point is significantly altered while the rest is fixed, the end result of the individual updates will drastically differ from that of the simultaneous update.

The parallel heuristic~(Alg.~\ref{algo:parallel_heuristic} in Appx. Sec.~\ref{sec:algorithms}) takes as input a local update function. 
The local update function~(Alg.~\ref{algo:hill_climbing} in Appx. Sec.~\ref{sec:algorithms} ) checks the constraint of fixed outline of the placement region. 
We call feasible SPs those SPs that fit into the placement region.

By using Alg.~\ref{algo:hill_climbing}~(Appx. Sec.~\ref{sec:algorithms}) as the local update function for the parallel heuristic~(Alg.~\ref{algo:parallel_heuristic} in Appx. Sec.~\ref{sec:algorithms}), the latter collects feasible points by accumulating the feasible sets generated by the former.
When the local update function~(Alg.~\ref{algo:hill_climbing} in Appx. Sec.~\ref{sec:algorithms}) is invoked in the parallel heuristic~(Alg.~\ref{algo:parallel_heuristic} in Appx. Sec.~\ref{sec:algorithms}), $c_{feasible}(\cdot)$ is the function which asserts the fixed-outline constraint using the LCS algorithm, and $\calN(x)$ is the set of neighbors of the sequence pair $x$ obtained by swapping adjacent elements in each permutation.

Given the kernel $K_{sp}$, the batch acquisition function, and the parallel heuristic for its optimization, we present the complete  Bayesian optimization for macro placement workflow in Alg.~\ref{algo:bbo_macro} in Appx. Sec.~\ref{sec:algorithms}. 
See also Fig.~\ref{fig:bo_flow} for a graphical illustration.




\section{Experiments}\label{sec:experiments}

\begin{figure*}[htbp]
    \centering
    \includegraphics[width=.95\linewidth]{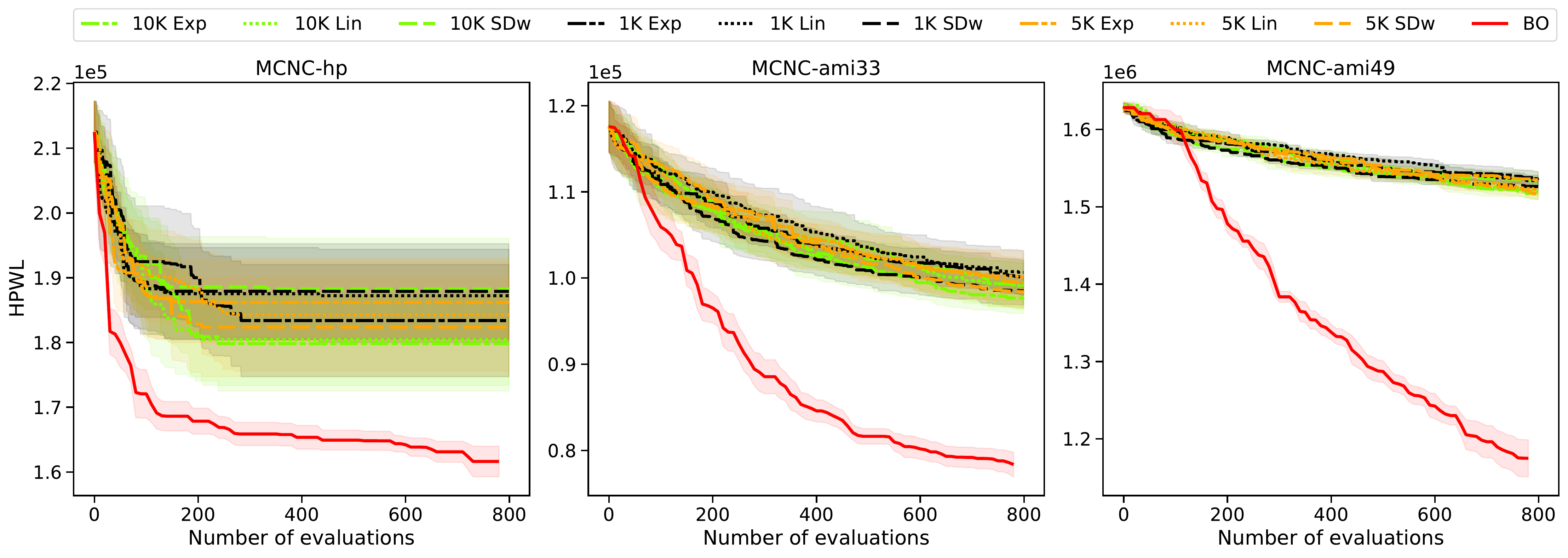}
    \caption{BO vs. SA on MCNC (\textit{hp}, \textit{ami33}, \textit{ami49}) across temperatures (1K, 5K, 10K) and schedules (linear, exponential, stepdown)}
    \label{fig:MCNC}
\end{figure*}


As a demonstration of the potential of BO, we test it on the MCNC benchmark \citep{mcnc}\footnote{\url{http://vlsicad.eecs.umich.edu/BK/MCNCbench/}} and present the results in Tab.~\ref{tab:MCNC}.
The optimization objective is to minimize HPWL which connects macros pins to I/O pads under the fixed-outline constraint.
Note that I/O pads are fixed on the boundary of the placement region. 
Under the relative location constraints specified by a sequence pair, we perform the linear constrained programming to minimize HPWL.
Note that this objective is simpler and cheaper-to-evaluate than the ones where BO can possibly show its full strengths.
Nonetheless, this experiment does indicate the potential of BO in macro placement.

\begin{table}[htbp]
\begin{threeparttable}
    \begin{center}
        \caption{HPWL results on the MCNC benchmark. (The number of macros in parentheses)\\ 
        }
        \label{tab:MCNC}
        \small
        \begin{tabular}{@{}rccccc@{}}
            \toprule
            \textbf{}& \textit{apte} (9)& \textit{xerox} (10) & \textit{hp} (11) & \textit{ami33} (33) & \textit{ami49} (49) \\
            \midrule
            eWL\citep{FUNKE2016355} & 513,061 & 370,993 & 153,328 & 58,627 & ~~640,509 \\
            ELS~\citep{liu2008net} & 614,602 & 404,278 & 253,366 & 96,205 & 1,070,010\\
            FD$^a$~\citep{samaranayake2009development} & 545,136 & 755,410 & 155,463 & 63,125 & ~~871,128\\
            \midrule
            \multirow{3}{*}{SA$^{b,d}$} &  515,570 &    431,108 &    179,826 &     97,691 &   1,517,051\\
                                        & $\pm$525 & $\pm$15,312 & $\pm$6,550 & $\pm$1,592 & $\pm$5,095\\
                                        & \textit{1e3, Exp} &    \textit{1e3, Lin} &   \textit{1e4, Exp} &   \textit{1e4, Exp} &    \textit{1e4, Lin}\\
            \hline
            \multirow{2}{*}{BO$^{c,d}$} &  514,138 &    388,936 &    161,620 &     78,359 &   1,174,972\\
                                        & $\pm$264 & $\pm$3,700 & $\pm$2,113 & $\pm$1,271 & $\pm$21,396\\
            \bottomrule
        \end{tabular}
        \begin{tablenotes}[flushleft]\scriptsize
            \item[a] Packings for \textit{hp}, \textit{ami33}, and \textit{ami49} have overlaps.
            \item[b] Among different temperature scheduling, the result with the lowest mean is reported.
            \item[c] BO uses the batch size $B=10$
            \item[d] Mean and standard error of 5 runs are reported.
        \end{tablenotes}
    \end{center}
\end{threeparttable}
\end{table}

In Tab.~\ref{tab:MCNC}, we compare the black-box function optimizers~(BO, SA) with other methods in the literature on all 5 MCNC problems.
In Fig~\ref{fig:MCNC}, we compare BO and SA with different temperature scheduling for 3 largest problems.
For all problems, BO has superior performance for the same number of evaluations, with the gap growing larger for larger number of macros.
In comparison with the other methods of Tab.~\ref{tab:MCNC}, we can see that BO performs competitively with only 520 evaluations of the objective on \textit{apte} which is the smallest problem.
We acknowledge that on the designs with a larger number of macros, \textit{ami33} and \textit{ami49}, there is a non-negligible gap between BO and eWL~\citep{FUNKE2016355}.
However, we expect that this gap does not translate to the real world applications that we envision since eWL cannot optimize
macro placements with standard cells, while BO and SA can.
This is because eWL relies on efficient HPWL evaluation and uses a far higher number of evaluations.
Moreover, for \textit{apte}, \textit{xerox}, \textit{hp}, eWL performed an exhaustive search.
On the other hand, due to super-exponential size of the space of sequence pairs on \textit{ami33} and \textit{ami49}, evaluations are not performed exhaustively but nevertheless are many orders of magnitude larger than the number of BO evaluations.
In comparison with ELS~\citep{liu2008net}, BO outperforms in all but \textit{ami49}. 
However, ELS is a SA tuned for a specific proxy cost function and we expect it not to be transferable to optimize more general and expensive cost functions.
Further, in spite of extremely small number of evaluations compared with the size of the search space in \textit{ami49}, BO demonstrates its potential for more general and more realistic objectives on such large number of macros.
FD~\citep{samaranayake2009development} outperforms BO on \textit{hp}, \textit{ami33}, \textit{ami49} but the macro locations that FD outputs have overlaps, while our method does not.



\section{Conclusions}
In this paper, we demonstrated the effectiveness of Bayesian optimization for macro placement, and have shown that it performs competitively with exhaustive search techniques on small benchmarks and performs reasonably well within compute constraints for large benchmarks. 
In comparison to simulated annealing, we have shown our BO framework to outperform across benchmarks with fewer evaluations. 
As mentioned above, realistic macro placement quality evaluation requires an expensive global placement loop. 
Our optimization objective in the experiments was to minimize HPWL of macro to I/O pads connections which helped us evaluate macro placement quality without standard cell placement in the loop. 
In the future, we plan \textit{1)} to extend this framework with an objective that considers standard cell placement for HPWL computation and congestion estimation \textit{2)} to utilize the current work’s output as initial solution to macro placement with subsequent standard cell placement \textit{3)} to extend our constraints with memory stacking requirements, dataflow constraints, channel and snapping constraints, which are typical in industry standard IPs.
On the machine learning front, a future challenge is to develop a BO framework that transfers across designs.

\vskip 0.2in
\bibliography{main}

\begin{thebibliography}{45}
\providecommand{\natexlab}[1]{#1}
\providecommand{\url}[1]{\texttt{#1}}
\expandafter\ifx\csname urlstyle\endcsname\relax
  \providecommand{\doi}[1]{doi: #1}\else
  \providecommand{\doi}{doi: \begingroup \urlstyle{rm}\Url}\fi

\bibitem[ope(2021)]{openroad}
The-openroad-project.
\newblock \url{https://github.com/The-OpenROAD-Project/OpenROAD}, 2021.

\bibitem[Adya and Markov(2005)]{10.1145/1044111.1044116}
S.~N. Adya and Igor~L. Markov.
\newblock Combinatorial techniques for mixed-size placement.
\newblock \emph{ACM Trans. Des. Autom. Electron. Syst.}, 10\penalty0
  (1):\penalty0 58–90, January 2005.
\newblock ISSN 1084-4309.
\newblock \doi{10.1145/1044111.1044116}.
\newblock URL \url{https://doi.org/10.1145/1044111.1044116}.

\bibitem[Adya and Markov(2002)]{adya2002consistent}
Saurabh~N Adya and Igor~L Markov.
\newblock Consistent placement of macro-blocks using floorplanning and
  standard-cell placement.
\newblock In \emph{Proceedings of the 2002 International Symposium on Physical
  design}, pages 12--17, 2002.

\bibitem[Adya and Markov(2001)]{adya2001}
S.N. Adya and I.L. Markov.
\newblock Fixed-outline floorplanning through better local search.
\newblock In \emph{Proceedings 2001 IEEE International Conference on Computer
  Design: VLSI in Computers and Processors. ICCD 2001}, pages 328--334, 2001.
\newblock \doi{10.1109/ICCD.2001.955047}.

\bibitem[Agnesina et~al.(2020)Agnesina, Chang, and Lim]{Agnesina2020}
Anthony Agnesina, Kyungwook Chang, and Sung~Kyu Lim.
\newblock Vlsi placement parameter optimization using deep reinforcement
  learning.
\newblock In \emph{2020 IEEE/ACM International Conference On Computer Aided
  Design (ICCAD)}, pages 1--9, 2020.

\bibitem[Baptista and Poloczek(2018)]{baptista2018bayesian}
Ricardo Baptista and Matthias Poloczek.
\newblock Bayesian optimization of combinatorial structures.
\newblock In \emph{International Conference on Machine Learning}, pages
  462--471. PMLR, 2018.

\bibitem[Brochu et~al.(2010)Brochu, Cora, and De~Freitas]{brochu2010tutorial}
Eric Brochu, Vlad~M Cora, and Nando De~Freitas.
\newblock A tutorial on bayesian optimization of expensive cost functions, with
  application to active user modeling and hierarchical reinforcement learning.
\newblock \emph{arXiv preprint arXiv:1012.2599}, 2010.

\bibitem[Char et~al.(2019)Char, Chung, Neiswanger, Kandasamy, Nelson, Boyer,
  Kolemen, and Schneider]{char2019offline}
Ian Char, Youngseog Chung, Willie Neiswanger, Kirthevasan Kandasamy, Andrew~O
  Nelson, Mark Boyer, Egemen Kolemen, and Jeff Schneider.
\newblock Offline contextual bayesian optimization.
\newblock \emph{Advances in Neural Information Processing Systems},
  32:\penalty0 4627--4638, 2019.

\bibitem[Cheng et~al.(2019)Cheng, Kahng, Kang, and Wang]{replace}
Chung-Kuan Cheng, Andrew~B. Kahng, Ilgweon Kang, and Lutong Wang.
\newblock Replace: Advancing solution quality and routability validation in
  global placement.
\newblock \emph{IEEE Transactions on Computer-Aided Design of Integrated
  Circuits and Systems}, 38\penalty0 (9):\penalty0 1717--1730, 2019.
\newblock \doi{10.1109/TCAD.2018.2859220}.

\bibitem[Chowdhury and Gopalan(2017)]{chowdhury2017kernelized}
Sayak~Ray Chowdhury and Aditya Gopalan.
\newblock On kernelized multi-armed bandits.
\newblock In \emph{International Conference on Machine Learning}, pages
  844--853. PMLR, 2017.

\bibitem[Deshwal and Doppa(2021)]{deshwal2021combining}
Aryan Deshwal and Jana Doppa.
\newblock Combining latent space and structured kernels for bayesian
  optimization over combinatorial spaces.
\newblock \emph{Advances in Neural Information Processing Systems},
  34:\penalty0 8185--8200, 2021.

\bibitem[Deshwal et~al.(2021)Deshwal, Belakaria, and Doppa]{deshwal2021mercer}
Aryan Deshwal, Syrine Belakaria, and Janardhan~Rao Doppa.
\newblock Mercer features for efficient combinatorial bayesian optimization.
\newblock In \emph{Proceedings of the AAAI Conference on Artificial
  Intelligence}, volume~35, pages 7210--7218, 2021.

\bibitem[Deshwal et~al.(2022)Deshwal, Belakaria, Doppa, and
  Kim]{deshwal2022bayesian}
Aryan Deshwal, Syrine Belakaria, Janardhan~Rao Doppa, and Dae~Hyun Kim.
\newblock Bayesian optimization over permutation spaces.
\newblock In \emph{Proceedings of the AAAI Conference on Artificial
  Intelligence}, volume~36, pages 6515--6523, 2022.

\bibitem[Frazier(2018)]{frazier2018tutorial}
Peter~I Frazier.
\newblock A tutorial on bayesian optimization.
\newblock \emph{arXiv preprint arXiv:1807.02811}, 2018.

\bibitem[Funke et~al.(2016)Funke, Hougardy, and Schneider]{FUNKE2016355}
J.~Funke, S.~Hougardy, and J.~Schneider.
\newblock An exact algorithm for wirelength optimal placements in vlsi design.
\newblock \emph{Integration}, 52:\penalty0 355--366, 2016.
\newblock ISSN 0167-9260.
\newblock \doi{https://doi.org/10.1016/j.vlsi.2015.07.001}.
\newblock URL
  \url{https://www.sciencedirect.com/science/article/pii/S0167926015000802}.

\bibitem[Gong et~al.(2019)Gong, Peng, and Liu]{gong2019quantile}
Chengyue Gong, Jian Peng, and Qiang Liu.
\newblock Quantile stein variational gradient descent for batch bayesian
  optimization.
\newblock In \emph{International Conference on Machine Learning}, pages
  2347--2356. PMLR, 2019.

\bibitem[Gonz{\'a}lez et~al.(2016)Gonz{\'a}lez, Dai, Hennig, and
  Lawrence]{gonzalez2016batch}
Javier Gonz{\'a}lez, Zhenwen Dai, Philipp Hennig, and Neil Lawrence.
\newblock Batch bayesian optimization via local penalization.
\newblock In \emph{Artificial intelligence and statistics}, pages 648--657.
  PMLR, 2016.

\bibitem[Kahng et~al.(2011)Kahng, Lienig, Markov, and Hu]{kahng2011vlsi}
A.B. Kahng, J.~Lienig, I.L. Markov, and J.~Hu.
\newblock \emph{VLSI Physical Design: From Graph Partitioning to Timing
  Closure}.
\newblock Springer Netherlands, 2011.
\newblock ISBN 9789048195916.
\newblock URL \url{https://books.google.nl/books?id=DWUGHyFVpboC}.

\bibitem[Kahng(2018)]{kahng2018machine}
Andrew~B Kahng.
\newblock Machine learning applications in physical design: Recent results and
  directions.
\newblock In \emph{Proceedings of the 2018 International Symposium on Physical
  Design}, pages 68--73, 2018.

\bibitem[Kozminski(1991)]{mcnc}
K.~Kozminski.
\newblock Benchmarks for layout synthesis - evolution and current status.
\newblock In \emph{28th ACM/IEEE Design Automation Conference}, pages 265--270,
  1991.

\bibitem[Kulesza and Taskar(2011)]{kulesza2011k}
Alex Kulesza and Ben Taskar.
\newblock k-dpps: Fixed-size determinantal point processes.
\newblock In \emph{ICML}, 2011.

\bibitem[Kulesza and Taskar(2012)]{kulesza2012determinantal}
Alex Kulesza and Ben Taskar.
\newblock Determinantal point processes for machine learning.
\newblock \emph{arXiv preprint arXiv:1207.6083}, 2012.

\bibitem[Liu and Nannarelli(2008)]{liu2008net}
Wei Liu and Alberto Nannarelli.
\newblock Net balanced floorplanning based on elastic energy model.
\newblock In \emph{2008 NORCHIP}, pages 258--263. IEEE, 2008.

\bibitem[Lu et~al.(2015)Lu, Zhuang, Chen, Chang, Chang, Wong, Sha, Huang, Luo,
  Teng, and Cheng]{eplace-ms}
Jingwei Lu, Hao Zhuang, Pengwen Chen, Hongliang Chang, Chin-Chih Chang,
  Yiu-Chung Wong, Lu~Sha, Dennis Huang, Yufeng Luo, Chin-Chi Teng, and
  Chung-Kuan Cheng.
\newblock eplace-ms: Electrostatics-based placement for mixed-size circuits.
\newblock \emph{IEEE Transactions on Computer-Aided Design of Integrated
  Circuits and Systems}, 34\penalty0 (5):\penalty0 685--698, 2015.
\newblock \doi{10.1109/TCAD.2015.2391263}.

\bibitem[Lyu et~al.(2018)Lyu, Yang, Yan, Zhou, and Zeng]{lyu2018batch}
Wenlong Lyu, Fan Yang, Changhao Yan, Dian Zhou, and Xuan Zeng.
\newblock Batch bayesian optimization via multi-objective acquisition ensemble
  for automated analog circuit design.
\newblock In \emph{International conference on machine learning}, pages
  3306--3314. PMLR, 2018.

\bibitem[Markov et~al.(2015)Markov, Hu, and Kim]{markov_review}
Igor~L. Markov, Jin Hu, and Myung-Chul Kim.
\newblock Progress and challenges in vlsi placement research.
\newblock \emph{Proceedings of the IEEE}, 103\penalty0 (11):\penalty0
  1985--2003, 2015.
\newblock \doi{10.1109/JPROC.2015.2478963}.

\bibitem[Mirhoseini et~al.(2020)Mirhoseini, Goldie, Yazgan, Jiang, Songhori,
  Wang, Lee, Johnson, Pathak, Bae, Nazi, Pak, Tong, Srinivasa, Hang, Tuncer,
  Babu, Le, Laudon, Ho, Carpenter, and Dean]{mirhoseini2020chip}
Azalia Mirhoseini, Anna Goldie, Mustafa Yazgan, Joe Jiang, Ebrahim Songhori,
  Shen Wang, Young-Joon Lee, Eric Johnson, Omkar Pathak, Sungmin Bae, Azade
  Nazi, Jiwoo Pak, Andy Tong, Kavya Srinivasa, William Hang, Emre Tuncer, Anand
  Babu, Quoc~V. Le, James Laudon, Richard Ho, Roger Carpenter, and Jeff Dean.
\newblock Chip placement with deep reinforcement learning, 2020.

\bibitem[Murata et~al.(1996)Murata, Fujiyoshi, Nakatake, and
  Kajitani]{murata1996vlsi}
Hiroshi Murata, Kunihiro Fujiyoshi, Shigetoshi Nakatake, and Yoji Kajitani.
\newblock Vlsi module placement based on rectangle-packing by the
  sequence-pair.
\newblock \emph{IEEE Transactions on Computer-Aided Design of Integrated
  Circuits and Systems}, 15\penalty0 (12):\penalty0 1518--1524, 1996.

\bibitem[Nguyen et~al.(2021)Nguyen, Le, Yamada, and Osborne]{nguyen2021optimal}
Vu~Nguyen, Tam Le, Makoto Yamada, and Michael~A Osborne.
\newblock Optimal transport kernels for sequential and parallel neural
  architecture search.
\newblock In \emph{International Conference on Machine Learning}, pages
  8084--8095. PMLR, 2021.

\bibitem[Oh et~al.(2019)Oh, Tomczak, Gavves, and Welling]{oh2019combinatorial}
Changyong Oh, Jakub Tomczak, Efstratios Gavves, and Max Welling.
\newblock Combinatorial bayesian optimization using the graph cartesian
  product.
\newblock \emph{Advances in Neural Information Processing Systems}, 32, 2019.

\bibitem[Oh et~al.(2021)Oh, Bondesan, Gavves, and Welling]{oh2021batch}
Changyong Oh, Roberto Bondesan, Efstratios Gavves, and Max Welling.
\newblock Batch bayesian optimization on permutations using acquisition
  weighted kernels.
\newblock \emph{arXiv preprint arXiv:2102.13382}, 2021.

\bibitem[Rasmussen(2003)]{rasmussen2003gaussian}
Carl~Edward Rasmussen.
\newblock Gaussian processes in machine learning.
\newblock In \emph{Summer school on machine learning}, pages 63--71. Springer,
  2003.

\bibitem[Samaranayake et~al.(2009)Samaranayake, Ji, and
  Ainscough]{samaranayake2009development}
Meththa Samaranayake, Helen Ji, and John Ainscough.
\newblock Development of a force directed module placement tool.
\newblock In \emph{2009 Ph. D. Research in Microelectronics and Electronics},
  pages 152--155. IEEE, 2009.

\bibitem[Shahriari et~al.(2015)Shahriari, Swersky, Wang, Adams, and
  De~Freitas]{shahriari2015taking}
Bobak Shahriari, Kevin Swersky, Ziyu Wang, Ryan~P Adams, and Nando De~Freitas.
\newblock Taking the human out of the loop: A review of bayesian optimization.
\newblock \emph{Proceedings of the IEEE}, 104\penalty0 (1):\penalty0 148--175,
  2015.

\bibitem[Snoek et~al.(2012)Snoek, Larochelle, and Adams]{snoek2012practical}
Jasper Snoek, Hugo Larochelle, and Ryan~P Adams.
\newblock Practical bayesian optimization of machine learning algorithms.
\newblock \emph{Advances in neural information processing systems}, 25, 2012.

\bibitem[Snoek et~al.(2015)Snoek, Rippel, Swersky, Kiros, Satish, Sundaram,
  Patwary, Prabhat, and Adams]{snoek2015scalable}
Jasper Snoek, Oren Rippel, Kevin Swersky, Ryan Kiros, Nadathur Satish,
  Narayanan Sundaram, Mostofa Patwary, Mr~Prabhat, and Ryan Adams.
\newblock Scalable bayesian optimization using deep neural networks.
\newblock In \emph{International conference on machine learning}, pages
  2171--2180. PMLR, 2015.

\bibitem[Srinivas et~al.(2009)Srinivas, Krause, Kakade, and
  Seeger]{srinivas2009gaussian}
Niranjan Srinivas, Andreas Krause, Sham~M Kakade, and Matthias Seeger.
\newblock Gaussian process optimization in the bandit setting: No regret and
  experimental design.
\newblock \emph{arXiv preprint arXiv:0912.3995}, 2009.

\bibitem[Sutton and Barto(2018)]{sutton2018reinforcement}
Richard~S Sutton and Andrew~G Barto.
\newblock \emph{Reinforcement learning: An introduction}.
\newblock MIT press, 2018.

\bibitem[Tang and Wong(2001)]{tang2001fasta}
Xiaoping Tang and DF~Wong.
\newblock Fast-sp: A fast algorithm for block placement based on sequence pair.
\newblock In \emph{Proceedings of the 2001 Asia and South Pacific design
  automation conference}, pages 521--526, 2001.

\bibitem[Tang et~al.(2001)Tang, Tian, and Wong]{tang2001fastb}
Xiaoping Tang, Ruiqi Tian, and DF~Wong.
\newblock Fast evaluation of sequence pair in block placement by longest common
  subsequence computation.
\newblock \emph{IEEE Transactions on Computer-Aided Design of Integrated
  Circuits and Systems}, 20\penalty0 (12):\penalty0 1406--1413, 2001.

\bibitem[Vashisht et~al.(2020)Vashisht, Rampal, Liao, Lu, Shanbhag, Fallon, and
  Kara]{vashisht2020placement}
Dhruv Vashisht, Harshit Rampal, Haiguang Liao, Yang Lu, Devika Shanbhag, Elias
  Fallon, and Levent~Burak Kara.
\newblock Placement in integrated circuits using cyclic reinforcement learning
  and simulated annealing.
\newblock \emph{arXiv preprint arXiv:2011.07577}, 2020.

\bibitem[Wang et~al.(2016)Wang, Zhou, and Jegelka]{wang2016optimization}
Zi~Wang, Bolei Zhou, and Stefanie Jegelka.
\newblock Optimization as estimation with gaussian processes in bandit
  settings.
\newblock In \emph{Artificial Intelligence and Statistics}, pages 1022--1031.
  PMLR, 2016.

\bibitem[Wu and Frazier(2016)]{wu2016parallel}
Jian Wu and Peter Frazier.
\newblock The parallel knowledge gradient method for batch bayesian
  optimization.
\newblock \emph{Advances in Neural Information Processing Systems},
  29:\penalty0 3126--3134, 2016.

\bibitem[Xu et~al.(2017)Xu, Liu, Zhao, Yang, Luo, and Zhang]{xu2017}
Chang Xu, Gai Liu, Ritchie Zhao, Stephen Yang, Guojie Luo, and Zhiru Zhang.
\newblock A parallel bandit-based approach for autotuning fpga compilation.
\newblock In \emph{Proceedings of the 2017 ACM/SIGDA International Symposium on
  Field-Programmable Gate Arrays}, FPGA '17, page 157–166, New York, NY, USA,
  2017. Association for Computing Machinery.
\newblock ISBN 9781450343541.
\newblock \doi{10.1145/3020078.3021747}.
\newblock URL \url{https://doi.org/10.1145/3020078.3021747}.

\bibitem[Zaefferer et~al.(2014)Zaefferer, Stork, and
  Bartz-Beielstein]{zaefferer2014distance}
Martin Zaefferer, J{\"o}rg Stork, and Thomas Bartz-Beielstein.
\newblock Distance measures for permutations in combinatorial efficient global
  optimization.
\newblock In \emph{International Conference on Parallel Problem Solving from
  Nature}, pages 373--383. Springer, 2014.

\end{thebibliography}

\newpage
\appendix

\section{Related Work}
Sequential macro placers~\citep{markov_review,adya2002consistent, 10.1145/1044111.1044116} produce overlap-free placements for macros in four steps:
\begin{enumerate*}
    \item cluster standard cells into soft rectangles,
    \item run a floorplanner on the original (hard) macros and new soft rectangles,
    \item remove the soft rectangles,
    \item place standard cells with fixed macros.
\end{enumerate*}
The floorplanner of choice is typically based on SA over sequence pairs with the most popular implementation being Parquet~\citep{adya2001} which incorporates several heuristics to select new configurations.
Modern sequential workflows such as the Triton macro placer included in the OpenRoad project~\citep{openroad} use RePlace~\citep{replace} for standard cell placement.
Replace is a state-of-the-art academic analytical placer that uses an electrostatic analogy whereby cells and macros are modelled as charged objects with charges proportional to their areas, and their electrostatic equilibrium leads to a uniformly spread placement.
Performing joint macro and standard cell placement using RePlace
produces overlaps that must be later removed by a legalization step, as done in \citep{eplace-ms} that also uses SA.

In all the aforementioned workflows we can replace SA with our BO algorithm. SA requires many iterations to converge and does not scale when using realistic cost functions, which limits the choice of cost functions that designers can feasibly use for SA and thus may lead to important aspects of the problem being ignored.
Furthermore, SA requires the designer to carefully adjust parameters such as temperature schedule and acceptance probability to obtain good results -- though~\citep{vashisht2020placement} proposes an algorithm that learns to propose good initial values. In contrast, in BO the kernel hyperparameters can be tuned automatically by fitting the training data with gradient-based optimization. Nevertheless, acquisition function maximization in our combinatorial setting requires some tuning, see Sec.~\ref{sec:method}.

Various techniques other than SA have been proposed for floorplanning. In \citep{FUNKE2016355} an exact enumeration algorithm is applied to larger problems using a divide-and-conquer strategy. However, this method can only be applied to the half-perimeter wire length objective and not more realistic cost functions.
Similarly, \citep{liu2008net, samaranayake2009development} also use wire length proxy functions.

A closely-related work to ours uses RL for macro placement~\citep{mirhoseini2020chip}.
RL requires many training iterations to converge to a good policy, while BO is more data-efficient and is therefore more appealing when evaluating an expensive reward function.
In contrast to RL, BO does not learn to act in multiple situations, meaning that each new design requires optimization from scratch. 
BO can be seen as a simplified instance of RL where one takes a single action (instead of a sequence of actions) in a fixed state (i.e.  bandits) \citep{sutton2018reinforcement,srinivas2009gaussian,chowdhury2017kernelized}.
Other practical differences of our work and \citep{mirhoseini2020chip} are: 1) their RL agent places macros sequentially while we jointly place all macros as done in SA; 2) they discretize the macro positions on a fictitious grid 
while we work in the exact continuum optimization formulation with no overlap constraints.
In Sec.\ref{sec:experiments} we compared our results against our SA implementation and previously reported methods on the same benchmark dataset.
We leave benchmarking against  \citep{mirhoseini2020chip}  as future work.

Recently, BO was tested on similar but much smaller cases in~\citep{deshwal2022bayesian}.
Their focus is on proposing a new kernel on permutations and it is orthogonal to our focus on the batch acquisition to tackle super-exponential growth.
In contrast to~\citep{deshwal2022bayesian}, our experiments were conducted on much larger spaces i.e., 3$\sim$4 times more macros -- in terms of the size of search space, this makes huge difference due to super-exponential growth of permutation spaces -- and demonstrated the effectiveness of the batch acquisition.
We leave the search for the optimal combination of the kernel and the batch acquisition for macro placement as a future work.

Finally, we note that in \citep{xu2017} a bandit-based approach similar to BO has been applied to optimizing parameters of FPGA compilation. 
This work does not tackle the challenges of large combinatorial spaces in macro placement.


\section{Background}
\subsection{Sequence pair}\label{subsec:sp}
Sequence pairs~(SPs) were introduced in \citep{murata1996vlsi} as a combinatorial representation for macro packing problems. 
We recall that a macro is a rectangle with distinguished points called pins which may connect wires.
For macros $\{m_1,\cdots,m_N\}$, an SP is a pair of permutations of length $N$, one per spatial dimension, and specifies the relative location of each pair of macros.
The relationship between the four possible relative locations of macros $m_i$ and $m_j$ and SPs are explained in~Tab.~\ref{tab:sequence_pair}.
\begin{table}[h]
    \centering
    \caption{Relative location specified by sequence pair $(\pi, \pi')$}
    \label{tab:sequence_pair}
    \small
    \begin{tabular}{@{}ccl@{}}
        \toprule
        $\pi$ & $\pi'$ & Relative location of $i$ and $j$\\ 
        \midrule
        $(\cdots,i,\cdots,j,\cdots)$ & $(\cdots,i,\cdots,j,\cdots)$ & $i$ is to the left of $j$\\ 
        $(\cdots,j,\cdots,i,\cdots)$ & $(\cdots,j,\cdots,i,\cdots)$ & $i$ is to the right of $j$\\ 
        $(\cdots,i,\cdots,j,\cdots)$ & $(\cdots,j,\cdots,i,\cdots)$ & $i$ is below $j$\\ 
        $(\cdots,j,\cdots,i,\cdots)$ & $(\cdots,i,\cdots,j,\cdots)$ & $i$ is above $j$\\ 
        \bottomrule
    \end{tabular}
\end{table}

Traditionally, the SP representation has been used in macro placement for optimizing area and half perimeter wire length (HPWL) \citep{murata1996vlsi}.
HPWL is the half perimeter of the bounding box around a net (e.g. the red boxes in Fig.~\ref{fig:bo_flow}).
To convert the SP to a packed placement, an algorithm called the Longest Common Subsequence (LCS) is used. It ensures minimal area placement, where no further vertical or horizontal adjustment of any macro is possible~\citep{murata1996vlsi}.

Simulated annealing~(SA) is commonly used to search over the space of SPs by carefully-designed stochastic moves \citep{adya2001}.
The optimization objective is typically a linear combination of area and HPWL. 
The conversion from an SP to a placement is the main computational bottleneck.
Since SA requires several thousands of evaluations to find a good solution, a cheap proxy for the objective that relies on LCS is used in practice~\citep{adya2001}. 
Another direction of work focused on the efficient LCS implementations to handle this computational bottleneck~\citep{tang2001fasta,tang2001fastb}.
In contrast, our work aims to optimize a complex and expensive objective through a BO routine, while using LCS to assert whether an SP can be converted to a placement which fits within the fixed placement region.

\subsection{Bayesian optimization}
BO has been widely successful in optimizing expensive-to-evaluate objectives such as hyperparameter optimization~\citep{snoek2015scalable}, neural architecture search~\citep{nguyen2021optimal} and optimization of the tokamak control for nuclear fusion~\citep{char2019offline}. The superior sample efficiency of BO is attributed to two components, namely the surrogate model and the acquisition function.
The surrogate model is a probabilistic model that approximates the objective while measuring the uncertainty of its approximation.
This uncertainty plays a crucial role in the exploration-exploitation trade-off. For this reason, Gaussian processes~(GPs) are widely used due to their superior uncertainty quantification~\citep{rasmussen2003gaussian,snoek2012practical}.
Given a point $x$ in the search space, at the $t$-th iteration of BO, the predictive mean $\mu_t(x)$ and the predictive covariance $K^{(t)}(x,x')$ of the GP surrogate model are defined as
\begin{align}
    &\mu_t(x) = m_x + K_{x,\bfX^{(t)}}(K_{\bfX^{(t)},\bfX^{(t)}} + \sigma^2 I)^{-1} (\bfy^{(t)} - m_{\bfX^{(t)}}) \nonumber \\
    \label{eq:Kt}
    &K^{(t)}(x, x') \!=\! K_{x,x'} - K_{x,\bfX^{(t)}}(K_{\bfX^{(t)},\bfX^{(t)}}\! +\! \sigma^2 I)^{-1} K_{\bfX^{(t)},x'}
\end{align}
where $m_{\cdot}$ is the mean function, $K_{\cdot,\cdot}$ is the kernel (i.e. prior covariance function), $\sigma^2$ is the variance of the observational noise and $\bfX^{(t)}$ is the set of points evaluated so far.
The predictive variance is $\sigma^2_t(x) = K^{(t)}(x,x)$.

Using the GP predictive distribution conditioned on the evaluation dataset $(\bfX^{(t)}, \bfy^{(t)})$, the acquisition function $a(\cdot, \cdot)$ quantifies the chance that the evaluation of a point improves the GP optimization.
An acquisition function is based on the intuition that the predictive mean and the predictive variance can be used to make an informed guess about the usefulness of a point in the input space~\citep{shahriari2015taking}: 
\begin{equation}
\label{eq:a_single}
a^{(t)}(x) = a(\mu_t(x), \sigma^2_t(x))\,.    
\end{equation}

In general, the acquisition function value is higher at points where the predictive mean and the predictive variance are relatively high.
The argument of the maximum of the acquisition function $x^{(t)}_{opt}$ is evaluated under the true objective $y^{(t)} = f(x^{(t)}_{opt})$.
This new datapoint is then added to the evaluation dataset and the BO process is repeated.
\begin{equation}
    \bfX^{(t + 1)} = [\bfX^{(t)}; x^{(t)}_{opt}], \,\,\,\, \bfy^{(t+1)} = [\bfy^{(t)}; y^{(t)}]
\end{equation}

BO can be accelerated when computational resources permit parallel evaluation of the objective. In this case, the acquisition function is defined over multiple points so that its optimization yields multiple points whose evaluation can be parallelized.
\begin{equation}
    \{x^{(t)}_{opt,b}\}_{b=1}^B = \argmax_{x_1, \cdots, x_B} a^{(t)}_{batch}(x_1, \cdots, x_B)
\end{equation}
This is called batch BO.
Several works have proposed methods which use different batch acquisition functions\citep{gonzalez2016batch,wu2016parallel,lyu2018batch,gong2019quantile}.
For a detailed overview of BO, the reader is referred to~\citep{brochu2010tutorial,shahriari2015taking,frazier2018tutorial}.

\section{Algorithms}\label{sec:algorithms}

\begin{algorithm}
    \caption{Parallel heuristic} \label{algo:parallel_heuristic}
    \begin{algorithmic}[1]
        \REQUIRE $a$ : an acquisition function for a single point\\
                 $a_{batch}$ : a batch acquisition function \\
                 $u_{local}(g(\cdot), x)$ : a local update function \\
                 ~~~~~~~such that $g(u_{local}(g(\cdot), x)) \ge g(x)$(maximization) \\
                 $\bfX_{feasible}$ : a feasible set
        \STATE $x_{opt,1} = \argmax_{x \in \bfX_{feasible}} a(x)$
        \STATE Randomly choose $x_{0,b}$ for $b = 2, \cdots, B$ from $\bfX_{feasible}$
        \REPEAT
            \STATE $\bfX_{feasible,s} = \emptyset$
            \FOR[\textbf{Parallel}]{$b \in \{2, \cdots, B\}$}
                \STATE Update $x_{s,b}$ \\
                      $a_{s,b}(x) = a_{batch}(x_{opt,1}, x_{s,2}, \cdots, x_{s,b-1}, x, x_{s,b+1}, \cdots)$\\
                      $x_{s+1,b}, \bfX_{feasible,s,b} = u_{local}(a_{s,b}(\cdot), x_{s,b})$
                \STATE Collect feasible sets \\
                      $\bfX_{feasible,s} = \bfX_{feasible,s} \cup \bfX_{feasible,s,b}$
            \ENDFOR
            \STATE Expand the feasible set\\
                  $\bfX_{feasible} = \bfX_{feasible} \cup \bfX_{feasible,s}$
            \STATE Update step count $s = s + 1$
        \UNTIL{Convergence or other stopping criteria}
        \RETURN $(x_{opt,1}, x_{\cdot,2}, \cdots, x_{\cdot,B}), \bfX_{feasible}$
    \end{algorithmic}
\end{algorithm}

\begin{algorithm}
    \caption{Local update with feasibility check} 
    \label{algo:hill_climbing}
    \begin{algorithmic}[1]
        \REQUIRE $g(x)$ : an objective function \\
                 $x_{old}$ : an initial point \\
                 $c_{feasible}(x)$ : a function checking the feasibility\\
                 $\calN(x)$ : a function listing neighbors of $x$
        \STATE Find neighbors of $x_{old}$, $\calN(x_{old})$\\
        \STATE Compute feasibility\\
              $\calN_{feasible}(x_{old}) = \{x \in \calN(x_{old}) \,\vert\, c_{feasible}(x) ~\text{is true}\}$
        \STATE Move toward the best feasible neighbor\\
              $x_{new} = \opt_{x \in \calN_{feasible}(x_{old})} g(x)$
        \STATE Expand the feasible set\\
              $X_{feasible} = X_{feasible} \cup \calN_{feasible}(x_{old})$
        \RETURN $x_{new}, X_{feasible}$
    \end{algorithmic}
\end{algorithm}

\begin{algorithm}
    \caption{Batch Bayesian optimization macro placement}
    \label{algo:bbo_macro}
    \begin{algorithmic}[1]
        \REQUIRE $f$ : the optimization objective \\
                 $\bfX^{(0)}_{feasible}$ : an initial feasible set \\
                 $\bfX^{(0)}, \bfy^{(0)}$ : an initial evaluation dataset
        \REPEAT
            \STATE Fit the surrogate model on the data ($\bfX^{(t)}$, $\bfy^{(t)}$)\\
                   $\mu_t(x)$, $\sigma^2_t(x)$ ~~~~~~~[\textbf{Used in the acquisition function}] \\
            \STATE Optimize the acquisition function \\
                   by calling Alg.~\ref{algo:parallel_heuristic} with \\
                   - Local update fn.: Alg.~\ref{algo:hill_climbing} \\
                   - Feasible set: $\bfX^{(t)}_{feasible}$ \\
                   $(x^{(t)}_1, \cdots, x^{(t)}_B)$, $\bfX^{new}_{feasible} \leftarrow$  \textbf{Alg.~\ref{algo:parallel_heuristic}}
            \STATE Evaluate the objective at $(x^{(t)}_1, \cdots, x^{(t)}_B)$ in \textbf{parallel} \\
                   $y^{(t)}_1 = f(x^{(t)}_1), \cdots, y^{(t)}_B = f(x^{(t)}_B)$
            \STATE Expand the evaluation dataset \\
                   $\bfX^{(t + 1)} = [\bfX^{(t)}; x^{(t)}_1; \cdots; x^{(t)}_B]$\\
                   $\bfy^{(t+1)} = [\bfy^{(t)}; y^{(t)}_1; \cdots; y^{(t)}_B]$
            \STATE Expand the feasible set \\
                   $\bfX^{(t+1)}_{feasible} = \bfX^{(t)}_{feasible} \cup \bfX^{new}_{feasible}$
            \STATE Update BO round count $t = t + 1$
        \UNTIL{Computational budget is exhausted}
        \RETURN $\bfX^{(\cdot)}$, $\bfy^{(\cdot)}$
    \end{algorithmic}
\end{algorithm}

\end{document}